\documentclass{article}

\usepackage[main, final]{neurips_2026}

\usepackage[utf8]{inputenc}
\usepackage[T1]{fontenc}
\usepackage{hyperref}
\usepackage{url}
\usepackage{booktabs}
\usepackage{amsfonts}
\usepackage{microtype}
\usepackage{xcolor}
\usepackage{graphicx}
\usepackage{natbib}
\usepackage{amsmath}

\title{Attention Alignment Between Humans and Vision-Language Models}

\author{%
  \begin{tabular}[t]{c}
  Isaac R. Christian$^{1}$, Udith Haputhanthri$^{2}$,
  Declan Campbell$^{1}$, \\
  Samuel Nastase$^{3}$, Taylor Webb$^{1,4}$, Michael Graziano$^{1,4}$ \\[0.3em]
  {\small \normalfont $^{1}$Princeton Neuroscience Institute, Princeton University} \\
  {\small \normalfont $^{2}$Department of Electrical and Computer Engineering, Princeton University} \\
  {\small \normalfont $^{3}$Department of Psychology and Center for Computational Language Sciences,} \\
  {\small \normalfont University of Southern California} \\
  {\small \normalfont $^{4}$Department of Psychology, Princeton University} 
  \end{tabular}
}

\begin{document}

\maketitle

\begin{abstract}
Visual perception depends on top-down goals and bottom-up sensory mechanisms. Vision-language models (VLMs) implement both, allowing us to treat each component as a separable hypothesis about what drives where we look. Following that logic, we trained four VLMs, crossing CNN and ViT encoders with LSTM and Transformer decoders in a 2$\times$2 factorial design. Our four models shared a training distribution, a vocabulary, a supervision signal, and every hyperparameter; only the encoder and decoder families differed. We extracted each model's attention weights on 200 images, for two separate tasks (social captioning and general description), and compared those weights to fixations from 49 participants performing the same tasks. We found that both decoder and encoder architecture shaped alignment, but decoder choice dominated. LSTM decoders reached 81--89\% of the human upperbound and Transformer decoders 47--74\%, with decoder architecture explaining nearly five times more variance in alignment than encoder architecture ($\eta^2 = 0.43$ vs.\ $0.09$). Despite their alignment advantage, LSTM-decoder attention maps were spatially diffuse and minimally task-differentiated; Transformer decoders, in contrast, were weaker in alignment yet showed sharper spatial concentration and stronger task differentiation. Attention weights also predicted fMRI signals from the Natural Scenes Dataset. Somewhat surprisingly, attention weights predicted early visual, ventral, and lateral cortex better than parietal cortex, a region often associated with the control of attention. And, unlike behavioral results in which decoder choice was most influential, encoder type contributed most to neural alignment, with the CNN driving the effect across decoder types. This dissociation suggests bottom-up and top-down VLM components make separable contributions to behavioral and neural attention alignment. 
\end{abstract}

\section{Introduction}
\label{sec:intro}

Artificial neural networks do a good job predicting, or aligning with, visual perception in primates \citep{schrimpf2018brainscore}.
Two factors influence how aligned a network becomes: the experiences it receives during training and the architecture it is originally endowed with \citep{doerig2023neuroconnectionist}. 
Results so far suggest experience is the more important factor. Behaviorally, very different architectures match or exceed human performance when trained on massive datasets \citep{geirhos2021partial, dehghani2023scaling}. Neurally, the same is true, with very different architectures predicting ventral stream responses equally well \citep{conwell2024large, linsley2025better, storrs2021diverse}. Yet, despite the dramatic success of simply training with more data, architecture becomes a relevant factor when input statistics are properly controlled. In the most extreme case, with no training at all, CNNs align with visual cortex far better than ViTs \citep{kazemian2025convolutional}.

How artificial and biological vision align is primarily studied by briefly flashing images to a viewer. Brief presentations restrict processing to bottom-up attention, which prioritizes incoming information by its visual statistics \citep{itti2001computational}. Top-down attention, in contrast, describes how that incoming information is modified by prior expectations. In cortex, this takes the form of goal-driven signals from frontoparietal regions biasing sensory representations throughout the visual hierarchy \citep{buffalo2010backward, corbetta2002control}. These top-down signals may have their own alignment signatures, influencing where people look and how cortex responds in ways that bottom-up processing alone cannot account for.

We set out to test the contribution of both top-down and bottom-up attention in visual alignment. We use vision language models (VLMs), which offer a way to dissociate the contribution of each. Inside VLMs, encoders mimic the bottom-up attentional component that segments incoming information into perceivable objects. Decoders use that segmentation, along with previously generated information, to decide what text to output next. We pair two encoders, a CNN and a ViT, with two decoders, an LSTM and a Transformer, creating a 2$\times$2 factorial design. We train these models to caption images for two tasks and examine their attention on 200 test images: one task asks VLMs to describe the image (the Describe task) and another asks them to caption what a person in the image is paying attention to (the Social task). We compare the attention patterns extracted from each VLM to the eye gaze of 49 participants performing the same tasks. We also compare the extracted attention patterns to neural signals from the Natural Scenes Dataset. Because we train models from scratch, the training statistics are completely controlled across architectures. What differs is the specific encoder-decoder combination, and the attention patterns that arise from each during training.

Our main contributions are:
\begin{itemize}
\item We train four models on identical data with the same objective, crossing two encoders (CNN, ViT) with two decoders (LSTM, Transformer). This isolates each component's contribution to attention alignment. Decoder architecture is the dominant predictor behaviorally, explaining nearly five times as much variance as encoder architecture ($\eta^2 = 0.43$ vs.\ $0.09$).
\item Encoder-decoder combinations predict different properties of visual selection. LSTM decoders better approximate where humans look but not how sharply they focus. Transformer decoders show the reverse, better matching how sharply attention is distributed but not where gaze is directed.
\item We fit attention weights, rather than intermediate representations, to brain activity from the Natural Scenes Dataset. Attention weights predicted activity across the visual hierarchy and outperformed low-level statistical controls beyond early visual cortex. Unlike behavioral results, encoder type drove neural alignment.
\end{itemize}

\section{Related Work}
\label{sec:related}

\paragraph{Human and machine attention.}
Attention has been studied extensively in computational neuroscience
\citep{itti2001computational, lindsay2020attention} and implemented
widely in artificial neural networks \citep{brauwers2021general, nguyen2018attentive}.
We focus on studies comparing human and artificial neural network attention in the visual domain.

\paragraph{Attention alignment in vision encoders.}
Passive attention identifies visual features that drive the model's output, without changing how the model processes each input. Active attention, in contrast, operates within the forward pass, reweighting intermediate representations for each input.
In vision models without a language decoder, active attention selects different image regions than humans do \citep{lai2020understanding, linsley2018learning}. 
It also predicts human visual selectivity less well than gradient-based passive methods \citep{langlois2021passive}. 
Passive maps align more closely with humans but still overlap only partially with human attention \citep{vandyck2021comparing, ebrahimpour2019humans, muller2024humans}, and have
become less aligned with human attention as model accuracy has
increased \citep{fel2022harmonizing}.
Training objective also influences alignment.
Active attention in vision transformers resembles human gaze after
self-supervised training, but not after supervised classification
training of the same architecture \citep{yamamoto2025emergence}.

\paragraph{Attention alignment in multimodal decoders.}
Multimodal tasks such as image captioning and visual question answering
use both an encoder and a decoder.
Active attention in the decoder selects image regions at each step of
generation.
It improves task performance \citep{xu2015show, anderson2018bottom}, but
it matches human attention poorly \citep{das2016human, chen2020air, patro2018differential, sood2021vqa}, and in captioning it matches human attention worse than a
bottom-up saliency model \citep{he2019human}.
Multimodal training improves alignment with humans but still falls well
short of human patterns \citep{kewenig2023multimodality}.
Across this literature, the contribution of specific architectural
choices to alignment has not been isolated.
Whether the gap in alignment comes from the encoder, the decoder, or their
interaction remains unclear.

\section{Methods}
\label{sec:methods}

\subsection{Participants and Stimuli}
\label{sec:methods:participants}

All procedures were approved by Princeton University's Institutional Review Board.
Participants ($n = 49$; 27 in the Social task, 22 in the Describe task) were recruited from an undergraduate population and provided written informed consent.
Eye movements were recorded using an SR Research EyeLink 1000 Plus (2000~Hz) while
participants viewed 200 naturalistic images, sampled at random from the COCO 2017 validation set \citep{lin2014microsoft} conditioned on containing at least one visible person.
In the \textit{Social task}, participants selected one person in the image and verbally
described the location of that person's attention (e.g., \textit{``The man is paying
attention to the red truck''}).
In the \textit{Describe task}, participants described the overall content of the scene (e.g., \textit{``A busy street with a man, a red truck, and several parked cars''}).
Gaze samples were convolved with a Gaussian kernel ($\sigma = 33.3$~px; width 200~px),
weighted by fixation duration, and downsampled to $256{\times}256$ pixels to yield
per-trial fixation heatmaps.

\subsection{Models}
\label{sec:methods:models}

\paragraph{Architecture variants.}
We trained four image captioning models in a 2$\times$2 factorial design on Social and Describe tasks (Table~\ref{tab:architectures}).

\textit{CNN-LSTM.} A ResNet-101 \citep{he2016deep} encoder (ImageNet pretrained)
extracts $14{\times}14$ feature maps (2048-dim per location, 196 spatial
positions total) via adaptive average pooling. At each decoding step $t$, the
encoder features and the previous LSTM hidden state are projected into a shared
space, and Bahdanau additive attention \citep{bahdanau2015neural} is computed
over these contextualized feature vectors. The resulting attention-weighted image
representation is concatenated with the previous word and submitted
to a single-layer LSTM decoder that generates one caption word per step.

\textit{CNN-Transformer.} The same ResNet-101 encoder feeds a
four-layer Transformer decoder (8 heads, $d_\mathrm{model}=512$, FFN width
2048) via a Linear(2048$\to$512) projection. Each decoder layer applies causal self-attention over the tokens generated so far, followed by multi-head cross-attention, in which each token's query is matched against keys from all 196 image patch positions to produce attention weights. The final step is a feed-forward block applied to each token independently.

\textit{ViT-LSTM.} A ViT-B/16 \citep{dosovitskiy2021image} encoder (ImageNet
pretrained) produces 196 patch tokens (768-dim), projected to 512 dimensions.
The same LSTM + Bahdanau attention decoder as CNN-LSTM is used,
treating the 196 projected patch tokens as the spatial memory.

\textit{ViT-Transformer.} The same ViT-B/16 encoder as ViT-LSTM feeds the same
four-layer Transformer decoder as CNN-Transformer.

\begin{table}[h]
\centering
\caption{Architecture comparison. All models share a $14{\times}14 = 196$
spatial grid and 512-dimensional decoder. During encoder fine-tuning, about 42M encoder parameters were trainable in every model.}
\label{tab:architectures}
\footnotesize
\resizebox{\linewidth}{!}{%
\begin{tabular}{llll}
\toprule
Model & Encoder (params) & Decoder (params) & Attention type \\
\midrule
CNN-LSTM        & ResNet-101 (42.6M) & LSTM (20.5M)        & Additive soft    \\
CNN-Transformer & ResNet-101 (42.6M) & Transformer (27.6M) & Multi-head cross \\
ViT-LSTM        & ViT-B/16 (86.2M)   & LSTM (14.2M)        & Additive soft    \\
ViT-Transformer & ViT-B/16 (86.2M)   & Transformer (26.8M) & Multi-head cross \\
\bottomrule
\end{tabular}}
\end{table}

\subsection{Training Curriculum}
\label{sec:methods:training}

All four architecture variants followed the same three-phase training sequence.

\textit{Phase 1 --- Describe Pretrain (encoder frozen).}
Decoder trained from scratch on COCO 2014 training captions
($\sim$414k image-caption pairs); encoder held at ImageNet initialization.

\textit{Phase 2 --- Describe Finetune (encoder unfrozen).}
Initialized from the best Phase 1 checkpoint (by BLEU-4 on COCO validation).
For ResNet-101, layers res3--res5 were unfrozen; for ViT-B/16, the top 6
transformer blocks (layers 6--11) plus LayerNorm and patch embedding were unfrozen.

\textit{Phase 3 --- Social Finetune (encoder unfrozen).}
Initialized from the best Phase 2 checkpoint and fine-tuned on a lab-collected social captioning dataset, with the encoder unfrozen using the same unfreezing strategy as Phase 2. The lab-collected dataset consisted of a separate cohort of 17 participants each performing the social task on a unique set of 200 images, randomly selected from the COCO 2017 training set. This yielded 3{,}299 image--caption pairs. Describe-task analyses use the best Phase 2 checkpoint and Social-task analyses use the best Phase 3 checkpoint.

All four models shared the same hyperparameters: embedding and decoder dimension 512, dropout 0.5, the Adam optimizer (decoder learning rate $4 \times 10^{-4}$; encoder $10^{-4}$ when unfrozen), gradient clipping at 5.0, and batch size 32 (80 in Phase 1). The learning rate was halved after 8 epochs without BLEU-4 improvement on COCO validation, and training stopped after 20 epochs. Captions were generated by beam search ($k = 5$).

\subsection{Attention Extraction and Head Selection}
\label{sec:methods:attention}

For LSTM-decoder models, the Bahdanau mechanism produces a 196-dimensional
vector $\alpha_t$ at each decoding step; per-image maps are obtained by averaging over
all generated words, reshaping to $14{\times}14$, and bilinearly upsampling to
$256{\times}256$. For Transformer-decoder models, cross-attention weights are averaged over generation steps in the same way, separately for each of the 32 heads (4 layers $\times$ 8 heads). Every attention map was then normalized to sum to 1.

\paragraph{Head selection.}
For the Transformer decoders we report the head that best aligns with human fixations, chosen by five-fold cross-validation over 200 images, for each task separately. The same head was chosen on 5 of 5 folds for both Transformer decoders in both tasks, indicating that head selection was stable. Appendix~\ref{app:head-selection} gives the full procedure.

\subsection{Baselines}
\label{sec:methods:baselines}

To ask whether learned attention predicts fixation better than established models of visual salience, we scored three standard saliency models against our models on the same 200 images. Those saliency models were Itti-Koch \citep{itti1998model}, GBVS \citep{harel2006graph}, and spectral residual \citep{hou2007saliency}. Itti-Koch captures local contrast, marking locations whose intensity, color, or orientation contrasts with their immediate surroundings, at various spatial scales; GBVS captures global contrast, marking how distinct each location in the image is from others; and spectral residual captures statistical surprise, marking the parts of an image that depart from the typical frequency content of natural images. Each saliency map was blurred with a Gaussian at that model's best width and compared with participants' fixation heatmaps in the same way as the model attention maps.

We also scored two large pretrained VLMs, \textit{Molmo 7B-D} \citep{deitke2024molmo} and \textit{Qwen3.5 9B} \citep{qwen35team2025}, as baselines. Both were trained at scale for captioning and grounding but were not fine-tuned on our tasks. Their attention maps were extracted, selected by head, and scored in the same way as those of our Transformer decoders (Appendix~\ref{app:head-selection}).

\subsection{Alignment Metrics and Statistics}
\label{sec:methods:stats}

For each participant and image, we compute the Spearman rank correlation between the flattened model map and the participant's fixation heatmap (both $256{\times}256$), Fisher-$z$ transform it, average across images within participant, and report the back-transformed mean $M$ with the SEM across participants. The human upperbound is the mean pairwise participant--participant correlation computed the same way ($r = 0.272$ Social, $0.248$ Describe); percentages of the upperbound divide a model’s Fisher-z mean by that of the upperbound.

The spread of attention is the Shannon entropy of the $14{\times}14$ map, normalized by its maximum, so that $H = 1$ is uniform attention and $H = 0$ a single peak; $196^{H}$ gives the effective number of attended patches. Between-task similarity is the Spearman $r$ between a model's Social and Describe maps of the same image. For the Transformer decoders, both maps come from one head, whichever of its two task heads reached the higher percentage of the upperbound. The human reference is the Spearman $r$ between the group-mean Social and Describe maps of each image (maps rank-transformed per participant before averaging), Fisher-$z$ averaged over images ($M = 0.658$, $\mathrm{SEM} = 0.005$).

Every pairwise architecture contrast (e.g., CNN-LSTM vs ViT-LSTM) is tested with a paired sign-flip permutation test on the 200 per-image differences (alignment averaged over participants within image; 10,000 permutations), with Cohen's $d_z$ as the effect size and Holm correction across the four contrasts within each task \citep{holm1979simple}. As a secondary check, we fit a linear mixed model to the Fisher-$z$ values of the four trained models with encoder, decoder, and task as effect-coded fixed effects (all interactions), a random intercept per participant, and a random effect for each participant-by-image combination (38,956 observations, 49 participants). Variance explained by each factor is reported as $\eta^2$ on participant-level mean alignment.

\subsection{Neural Encoding Model}
\label{sec:methods:neural}

We asked whether model attention maps predict brain responses from the Natural Scenes Dataset (NSD; \citealp{allen2022massive}). We used NSD's shared images (907 to 1,000 per participant), surface-projected single-trial betas (\texttt{betas\_fithrf\_GLMdenoise\_RR}, fsaverage5) $z$-scored within session and averaged over the three repetitions of each image, and restricted analysis to the 5,032 vertices with mean noise-ceiling SNR above 0.2 in every participant.

For each participant we fit a ridge regression predicting every vertex from the model's normalized 196-dimensional attention map for each image, with the penalty selected per vertex by nested cross-validation (5 outer folds, 4 inner folds, $\alpha \in 10^{1}$--$10^{10}$). Sobel edge energy and mean luminance in each of the 196 cells were fit the same way, serving as low-level controls. We averaged the per-vertex $R^2$ within four regions of interest spanning the visual hierarchy (early visual, ventral, lateral, and parietal), defined by grouping vertices according to NSD's \texttt{streams} atlas.

Each predictor is referenced to its own permutation floor, obtained by refitting that predictor's maps with the image labels shuffled. A predictor counts as informative when its $R^2$ exceeds its floor, tested with a paired $t$-test across the 8 participants and Holm-corrected within predictor across the four regions.

Architecture and region effects were tested with a repeated-measures ANOVA on the floor-corrected values, with encoder, decoder, region, and task as within-participant factors and Greenhouse-Geisser correction applied to effects with more than two levels. Simple effects average each model across the four regions within participant. Planned contrasts are paired $t$-tests across participants, Holm-corrected within each family of comparisons (region, low-level control, architecture).

\section{Results}
\label{sec:results}

\subsection{Model-Human Alignment}
\label{sec:results:alignment}

\subsubsection{Decoder Architecture Drives Behavioral Alignment}
\label{sec:results:alignment:decoder}

\begin{figure}[t]
 \centering
 \includegraphics[width=\linewidth]{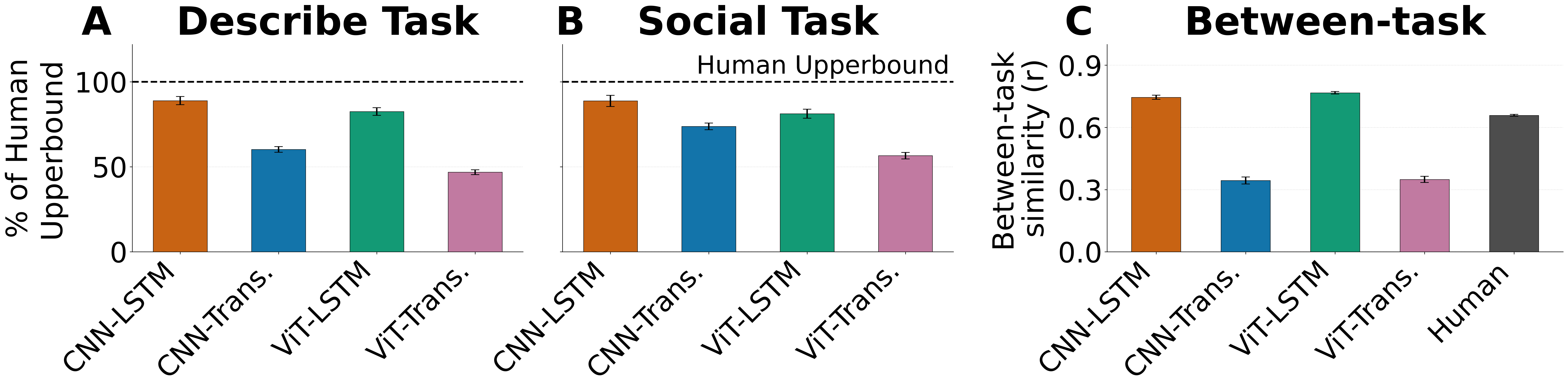}
\caption{\textbf{(A)} Describe task alignment. Model attention maps were correlated with human fixation patterns, and correlations were averaged across images and participants. \textbf{(B)} Same as A for the Social task. \textbf{(C)} Between-task spatial alignment, computed as the correlation between each agent's attention maps on the Social and Describe versions of each image. The human bar is the correlation between the group-mean Social and group-mean Describe maps. Bars in (A--B) show \% of the human upperbound (dashed line).}
 \label{fig:alignment}
\end{figure}

We first tested which decoder architectures best matched human attention, holding the encoder constant. With a CNN encoder, the LSTM decoder was more aligned than the Transformer decoder on both tasks: CNN-LSTM reached 88.8\% and 89.0\% of the upperbound on the Social and Describe tasks, respectively ($M_\mathrm{soc} = 0.243$, $M_\mathrm{des} = 0.222$), while CNN-Transformer reached 73.8\% and 60.3\% ($M_\mathrm{soc} = 0.203$, $M_\mathrm{des} = 0.152$; Social $d_z = 0.70$, Describe $d_z = 0.83$, both $p_\mathrm{Holm} < .001$). The same pattern was observed when the ViT encoder was fixed: ViT-LSTM reached 81.3\% and 82.5\% of the upperbound ($M_\mathrm{soc} = 0.223$, $M_\mathrm{des} = 0.206$), while ViT-Transformer reached 56.6\% and 46.8\% ($M_\mathrm{soc} = 0.157$, $M_\mathrm{des} = 0.118$; $d_z = 1.08$ and $1.18$, both $p_\mathrm{Holm} < .001$). Within each encoder family and in both tasks, the LSTM decoder was more aligned than the Transformer decoder (Figure~\ref{fig:alignment}). This ordering holds across five additional alignment metrics (Appendix~\ref{app:metric-suite}) and at each model's best Gaussian blur (Appendix~\ref{app:smoothness}).

With the decoder held constant, we tested which encoder architecture contributed most to alignment. Within each decoder family, CNN encoders outperformed ViT encoders by 6--17 percentage points (LSTM: $d_z = 0.50$ Social, $0.29$ Describe; Transformer: $d_z = 0.64$ Social, $0.35$ Describe; all $p_\mathrm{Holm} < .001$). Just as one decoder architecture (the LSTM) was more aligned across encoder families, one encoder architecture (the CNN) was more aligned across decoder families.

In comparing the effect of decoder versus encoder, we found decoder architecture was the stronger driver of alignment, accounting for $\eta^2 = 0.43$ of participant-level variance in alignment against $\eta^2 = 0.09$ for the encoder. The mixed model gave the same picture: the decoder main effect (Transformer vs.\ LSTM, $\beta = -0.069$, $z = -107.1$) was more than twice the encoder main effect (ViT vs.\ CNN, $\beta = -0.030$, $z = -46.7$), and the two interacted, a ViT encoder reducing alignment more with a Transformer decoder than with an LSTM ($\beta = -0.022$, $z = -17.5$). The decoder effect was smaller in the Social task than in the Describe task ($\beta = 0.026$, $z = 20.5$; all $p < .001$).

\subsubsection{Attention Adaptation Across Tasks}
\label{sec:results:flexibility}

\begin{figure}[t]
  \centering
  \includegraphics[width=\linewidth]{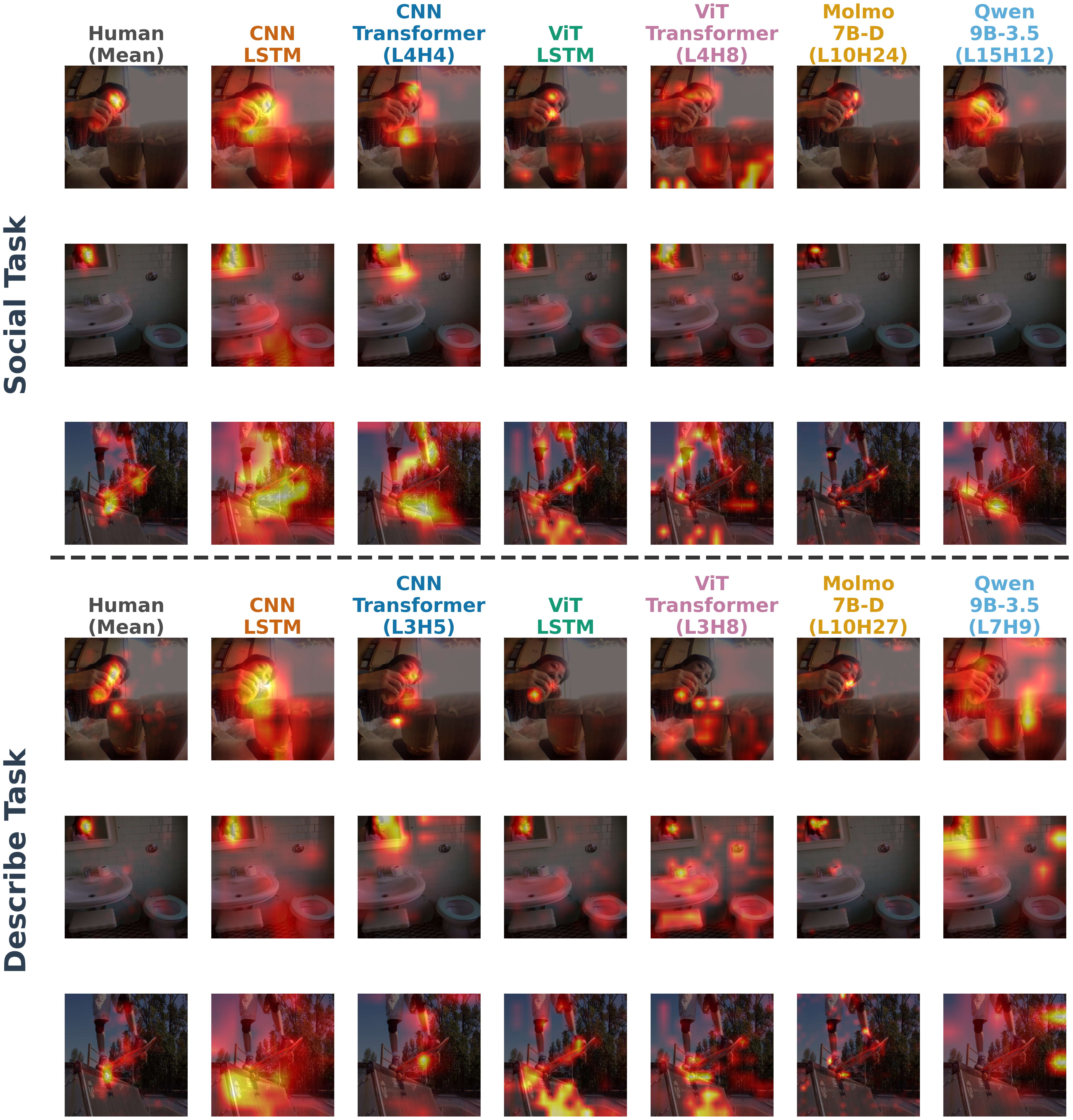}
  \caption{Attention map examples on three images and two tasks. Top rows: Social task. Bottom rows: Describe task. Columns: Human gaze patterns (mean), CNN-LSTM, CNN-Transformer, ViT-LSTM, ViT-Transformer, Molmo 7B-D, and Qwen3.5 9B. Molmo and Qwen3.5, VLMs trained at scale, are shown for comparison; their alignment scores are in Appendix~\ref{app:baselines}.}
  \label{fig:flexibility}
\end{figure}

Next, we asked whether models shift their attention across tasks as humans do. As visualized in Figure~\ref{fig:flexibility}, participants attend differently when describing a general scene versus the location of someone's attention. We expected that models with more human-like attention would adapt attention between tasks in a manner comparable to humans.

Despite showing the strongest alignment with human fixations (Figures~\ref{fig:alignment}A--B), LSTM-decoder models adapted less across tasks. Both CNN-LSTM ($M = 0.745$, $\mathrm{SEM} = 0.010$) and ViT-LSTM ($M = 0.767$, $\mathrm{SEM} = 0.006$) exceeded the human reference ($M = 0.658$), meaning their attention maps were more consistent across tasks than human fixations were (Figure~\ref{fig:alignment}C). CNN-Transformer ($M = 0.344$, $\mathrm{SEM} = 0.017$) and ViT-Transformer ($M = 0.349$, $\mathrm{SEM} = 0.015$) both fell well below the human level, showing the most task-differentiated attention of the four architectures.

\subsubsection{Spread of Spatial Attention}
\label{sec:results:entropy}

Fixation alignment captures which regions models and participants prioritized, but not how they distributed attention within each image. To characterize the spread of attention, we computed normalized Shannon entropy of each model's attention over the $14{\times}14$ patch grid (196 patches), where $H = 1$ corresponds to attention distributed equally across all image patches and $H = 0$ to a single concentrated peak.

Human fixation heatmaps were sharply concentrated ($H = 0.552$ Social, $H = 0.617$ Describe), equivalent to attention focusing on roughly 18 and 26 of the 196 image patches, on average. In contrast, LSTM-decoder models were nearly maximally diffuse across both tasks (CNN-LSTM and ViT-LSTM: $H \approx 0.999$), with some attention on nearly all patches (${\sim}$195). Transformer-decoder models fell between LSTM models and humans (CNN-Transformer: $H = 0.855$--$0.904$, ${\sim}$91--118 patches; ViT-Transformer: $H = 0.908$--$0.946$, ${\sim}$121--148 patches). These results separate two aspects of alignment with human attention. LSTM decoders better approximate where humans look but not how sharply they focus. Transformer decoders show the reverse, better matching how sharply attention is distributed but less so where gaze is directed.

\subsection{Baselines}
\label{sec:results:baselines}

We compared each model's learned attention to established predictors of visual attention patterns. We focused on three standard measures of saliency: Itti-Koch, GBVS, and spectral residual. All three define where attention should be placed according to visual statistics of the image. We ran these predictions through each of our 200 images and compared their alignment to model alignment. The most aligned model, the CNN-LSTM, beat all three saliency measures in the Social task ($d_z = 0.45$ to $1.25$, all $p < .001$), and beat all but GBVS in the Describe task ($d_z = 0.32$ and $0.99$, both $p < .001$). The GBVS difference was not significant ($d_z = -0.12$, $p = .09$). Our best model, which includes both top-down and bottom-up components, accounts for information that strictly bottom-up models do not capture. This same model also showed comparable alignment scores to two large pretrained VLMs, Molmo 7B-D and Qwen3.5 9B, despite having about 63M parameters against their 7 to 9 billion (Figure~\ref{fig:flexibility}; Appendix~\ref{app:baselines}).

\subsection{Neural Encoding of Attention Maps}
\label{sec:results:neural}

\begin{figure}[t]
  \centering
  \includegraphics[width=\linewidth]{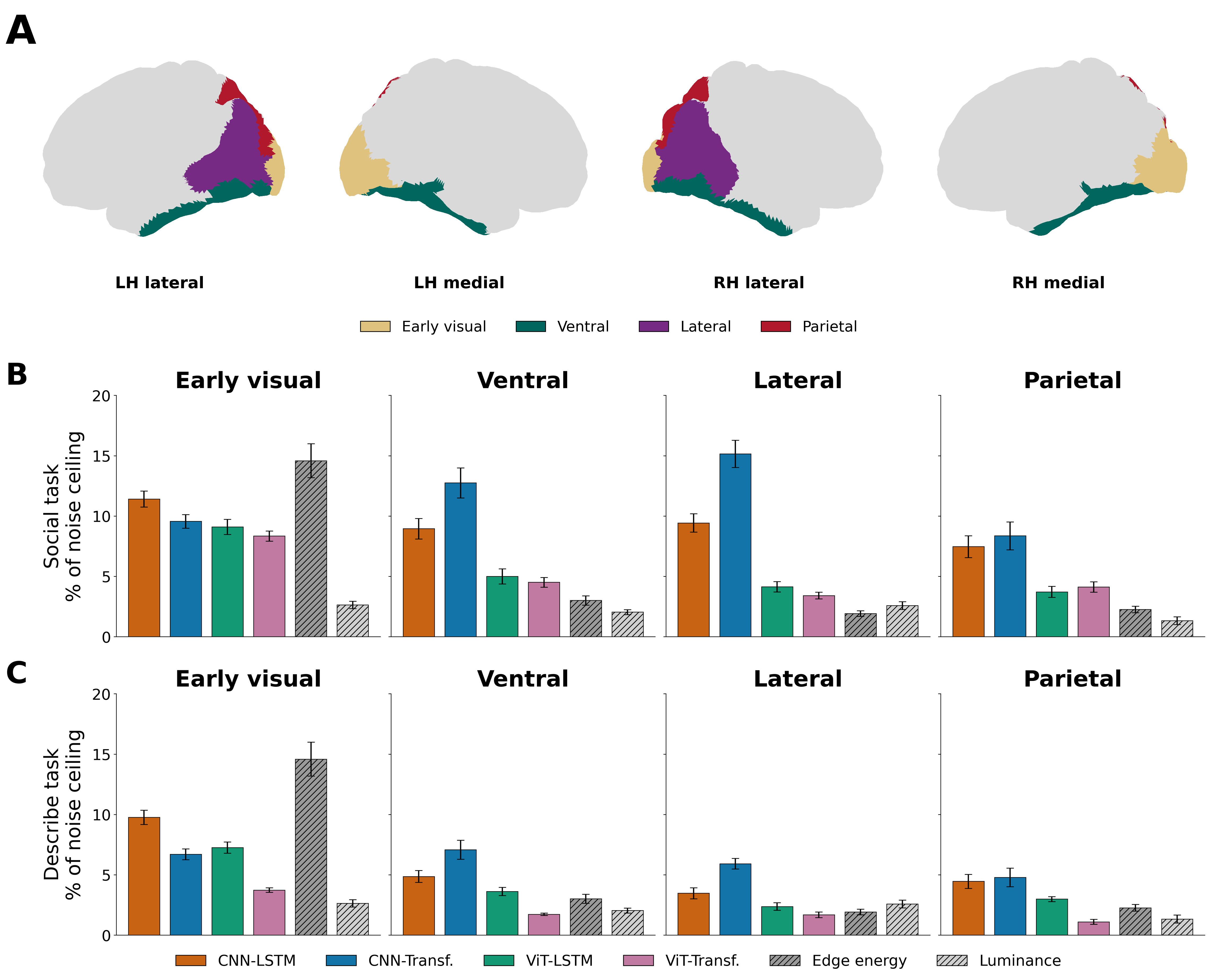}
  \caption{\textbf{(A)} The four regions of interest on the fsaverage5 surface, grouped from NSD's \texttt{streams} atlas (lateral and medial views of each hemisphere). \textbf{(B, C)} Variance in responses explained by each predictor's attention map (held-out test $R^2$) in the Social (B) and Describe (C) tasks, as a percentage of each participant's noise ceiling in that region. The ceiling is NSD's within-participant ceiling for responses averaged over three repetitions, averaged over the region's analyzed vertices. Every model and both low-level controls (Sobel edge energy, luminance) exceeded their own floor in every region in both tasks (all $p_\mathrm{Holm} < .05$). Error bars: SEM across the 8 participants.}
  \label{fig:neural}
\end{figure}

Previous analyses measured behavioral gaze alignment. Here we asked whether the same attention maps predicted neural responses, and if so, where. We had specific hypotheses about which regions attention weights would predict. We expected attention weights to be most aligned with activity in parietal cortex, an area involved in assigning spatial priority, or determining which areas of the visual field are worthy of attention \citep{behrmann2004parietal, bisley2010attention, silver2005topographic}. In contrast, we expected regions associated with other aspects of visual processing, such as contrast sensitivity or object recognition, to be less well predicted by attention weights. 

To test this hypothesis, we extracted BOLD activation in response to ${\sim}1{,}000$ images across 8 participants in the Natural Scenes Dataset. We selected four ROIs spanning the visual hierarchy (Early Visual, Ventral, Lateral, Parietal; Figure~\ref{fig:neural}A), regressing each ROI’s responses onto each model’s attention maps. We added two additional regressors, luminance and edge energy, which measured our models against the predictive power of low-level visual statistics. The addition of these regressors enabled a second hypothesis. We expected low level features would best predict early visual processing but fail at later stages, while models would be worse at capturing early and better at capturing later. We show results for trained models in Figure~\ref{fig:neural}; Qwen3.5, Molmo, and saliency model encoding results are available in Appendix~\ref{app:nsd-robustness}. 

As our first hypothesis predicted, we found that attention alignment varied by region ($F(3,21) = 66.94$, $p_\mathrm{GG} < .001$; Figure~\ref{fig:neural}B). However, it did not vary in the way we expected. In the parietal region, where attention maps were expected to be most predictive, we found that prediction was smallest here, relative to other ROIs, across all models (below early visual $d_z = 5.11$, below ventral $d_z = 2.52$, and below lateral $d_z = 1.81$, all $p < .01$). Model attention predicted early visual, lateral, and ventral areas better than the one canonically associated with spatial priority, contrary to our original hypothesis. 

Our second hypothesis held. In early visual cortex, edge energy was the strongest predictor relative to trained models ($d_z = 1.32$, $p < .01$). As we went up the visual hierarchy, however, low level controls lost their predictive power, beaten by models whose maps encompassed more than low level statistics (mean of the four trained models over edge energy: ventral $d_z = 1.94$ and lateral $d_z = 2.22$, both $p < .01$, parietal $d_z = 1.47$, $p < .01$; over luminance: ventral $d_z = 2.60$ and lateral $d_z = 2.41$, both $p < .01$, parietal $d_z = 1.96$, $p < .01$). We take this as evidence that trained attention maps encode information beyond low-level statistics, and that this information becomes more useful at later stages in visual processing.

As a final test, we examined whether encoder or decoder was the stronger driver of attention alignment across all regions of interest and both tasks. In behavioral results, we found choice of decoder was the largest contributor to alignment. The neural results showed the opposite pattern, with the encoder driving the effect instead (encoder: $F(1,7) = 43.45$, $p < .001$; decoder: $F(1,7) = 0.03$, $p = .86$). The CNN was the better encoder, outperforming the ViT under each decoder class (LSTM $d_z = 2.21$; Transformer $d_z = 2.36$; both $p < .01$). Additionally, which decoder was most aligned depended on the encoder it was paired with ($F(1,7) = 41.24$, $p < .001$). With a CNN encoder, the Transformer decoder predicted better than the LSTM ($d_z = 1.27$, $p < .01$), while with a ViT encoder, the LSTM decoder predicted better than the Transformer ($d_z = 1.56$, $p < .01$).  The neural and behavioral results, considered together, reveal a dissociation. Behaviorally, CNN encoders and LSTM decoders each improved gaze alignment, with the decoder contributing most. Neurally, the encoder carried the effect, with the CNN encoder more aligned regardless of its decoder.

\section{Discussion}
\label{sec:discussion}

In this study we compared machine and human attention alignment behaviorally and
neurally. Where the predominant approach focuses on bottom-up aspects of alignment
\citep{schrimpf2018brainscore}, we compared the relative contributions of bottom-up (encoder) and top-down (decoder) components 
to behavioral and neural alignment when all else (e.g., training data) was held constant. Rather
than constructing attention maps from the encoder 
\citep{langlois2021passive, ebrahimpour2019humans}, we compared human attention with the decoder’s attention over image regions. We found that which decoder type was
implemented -- either an LSTM or Transformer -- was more important for behavioral alignment,
while encoder type -- CNN or ViT -- was more important for neural alignment.

Top-down signals are one of several mechanisms proposed to make neural network models of vision more biologically plausible (e.g., topological loss: \citealp{margalit2024unifying}; developmentally inspired curriculum: \citealp{orhan2024learning, ayzenberg2026advancing}; foveation: \citealp{deza2020emergent}). Our models implement two forms of top-down signaling. In the first, previously generated text applies a gain signal to image features, much as spatial attention scales the responses of visual neurons \citep{mcadams1999effects, lindsay2020attention}. The two decoders compute this gain differently. The Transformer computes each weight as the similarity between learned projections of the text and image embeddings. The LSTM instead is free to learn what is relevant in a mixed visuo-semantic embedding. We compared these mechanisms directly, rather than asking whether either one aligns with humans on its own, which they often do not \citep{linsley2018learning, das2016human, he2019human}. Each mechanism introduced its own selection bias. LSTM attention was diffuse, changed little between tasks, and best matched where people looked. Transformer attention was sharper and changed more between tasks, as human attention did, but matched where people looked less well.

The second form changes the representations themselves rather than selecting among fixed
ones. During fine-tuning, we unfreeze encoder layers so gradients from the
captioning objective can change which features are extracted. This resembles cortical signals
descending from frontoparietal regions to modulate firing in visual cortex
\citep{luck1997neural,buffalo2010backward}, but differs in two ways. First, cortical feedback
operates within a trial, at the timescale of the percept, whereas ours operates across
training and is fixed at inference. Second, cortical feedback for spatial attention is
specific to the attended location, whereas gradient-based tuning changes feature selectivity
across the whole image.

Top-down components had clear effects, but so did bottom-up components. CNN encoders
outperformed ViTs, according to both behavioral and neural alignment scores. Despite many architectures aligning comparably with neural data when trained extensively \citep{conwell2024large, hosseini2024universality}, CNNs may hold an advantage over ViTs when training curriculum is controlled. We extend this result from untrained networks \citep{kazemian2025convolutional} to networks trained for image captioning, and show that although ViTs better match some human behavior in unimodal contexts (e.g., error consistency: \citealp{tuli2021convolutional}; shape bias: \citealp{naseer2021intriguing}), CNNs align better with human gaze in a multimodal context.

We were somewhat surprised that attention weights predicted ventral regions better than dorsal.
Whereas ventral regions encode semantic content \citep{ungerleider1982two},
dorsal parietal regions are associated with building a map of attentional priority
\citep{behrmann2004parietal, bisley2010attention}. The attention patterns that emerge from our models are exactly
this; contentless representations weighing the importance of spatial information. Model
attention patterns should align well here, if parietal regions are indeed mapping visual
priority. We found that while predictive, attention weights were more aligned with lateral,
ventral, and early visual ROIs. One explanation is that the task in our study (captioning
images) does not match the one performed by NSD participants. In
NSD, participants' gaze is fixed at the center of the scene while they indicate whether the
image was viewed before. If participants were instead asked to caption images and saccade freely,
as our participants did, parietal regions may align better with attention weights.

\paragraph{Limitations.}
Models and participants did not view the images in
the same way. When given a new image, people visually sample the scene to discover what is worth
captioning in the first place. A model does not explore this way, instead reading all 196
image locations at once. Diffuse maps may therefore align better with human maps, because by covering more of the image they can account for these exploratory fixations.

Our architecture comparison is not fully controlled. Parameter counts and depth
differ between the LSTM and Transformer decoders, so decoder type is entangled with
decoder size. Matching decoders
on FLOPs, or sweeping depth and width while holding the attention mechanism fixed, would
confirm these differences are specifically attributable to architecture.

\begin{ack}
We thank Hanna Hornfeld and Lauren Kim for their help with data collection and data analysis. I.R.C. was supported by the National Institute of Mental Health of the National Institutes of Health under award T32MH065214, the NRSA Training Grant in Quantitative Neuroscience at the Princeton Neuroscience Institute. The content is solely the responsibility of the authors and does not necessarily represent the official views of the National Institutes of Health. The authors declare no competing interests.
\end{ack}

\bibliographystyle{plainnat}
\bibliography{refs}

\appendix

\section*{Code and Data Availability}
Training and analysis code and the eye-tracking fixation heatmaps are available at \url{https://github.com/isaacrc/vlm-attention-alignment}. Trained model checkpoints are available at \url{https://huggingface.co/isaacrc/vlm-attention-alignment}.

\section{Head Selection}
\label{app:head-selection}

\paragraph{Cross-validated selection.}
For every model with more than one candidate map, per-image, participant-averaged Fisher-$z$ alignment was computed for every head. Within each task the 200 images were split into 5 folds; the head with the highest mean alignment on the 4 training folds was scored on the held-out fold, and the 5 held-out scores were averaged. Fold agreement was 5/5 for CNN-Transformer and ViT-Transformer in both tasks, for Qwen3.5 in the Social task, and for Molmo in the Describe task; 4/5 for Qwen3.5 in the Describe task; and 3/5 for Molmo in the Social task, where its two best heads (L10H24 and L10H27) differ by less than $r = 0.001$ over all 200 images. Winner's-curse inflation, the difference between the expected in-sample maximum (bootstrap over images, 2,000 resamples) and the cross-validated estimate, was $r \le 0.003$ in every model-by-task cell (largest for Qwen3.5 in the Describe task).

\paragraph{Baseline VLMs.}
For Molmo 7B-D and Qwen3.5 9B, attention was extracted for every head of every layer (Molmo: 28 layers $\times$ 28 heads; Qwen3.5: its 8 full-attention layers $\times$ 16 heads) and averaged over generated tokens. Each map was bilinearly upsampled from the model's native token grid to $256{\times}256$, the same interpolation as for the trained models, and normalized to sum to 1. Molmo's grid is $28{\times}28$, reassembled from its $3{\times}3$ overlapping image crops with the overlap margins, which carry no image tokens, removed first; Qwen3.5's grid is $8{\times}8$. One head per task was selected by the cross-validation above.

\section{Baselines}
\label{app:baselines}

We compared our models with three kinds of baselines. The first kind predicted where attention should land according to the visual statistics of each image. We used three established saliency models (Section~\ref{sec:methods:baselines}) as well as Sobel edge energy and luminance, each of which uniquely defines what is statistically important. The second baseline captured biases in where people tend to look. For instance, we know human attention is drawn to the center of images, to people, and to faces. We built three priors to represent these tendencies (center prior: the mean fixation map of all participants on the other 199 images; the face prior: a Gaussian blob on every annotated face; the person prior: the union of annotated person masks). The third baseline showed how our model's attention patterns stacked up against the attention of two VLMs trained at scale, Molmo 7B-D and Qwen3.5 9B. Each baseline was scored against human fixations in the same way as our models were.

\begin{figure}[h]
  \centering
  \includegraphics[width=\linewidth]{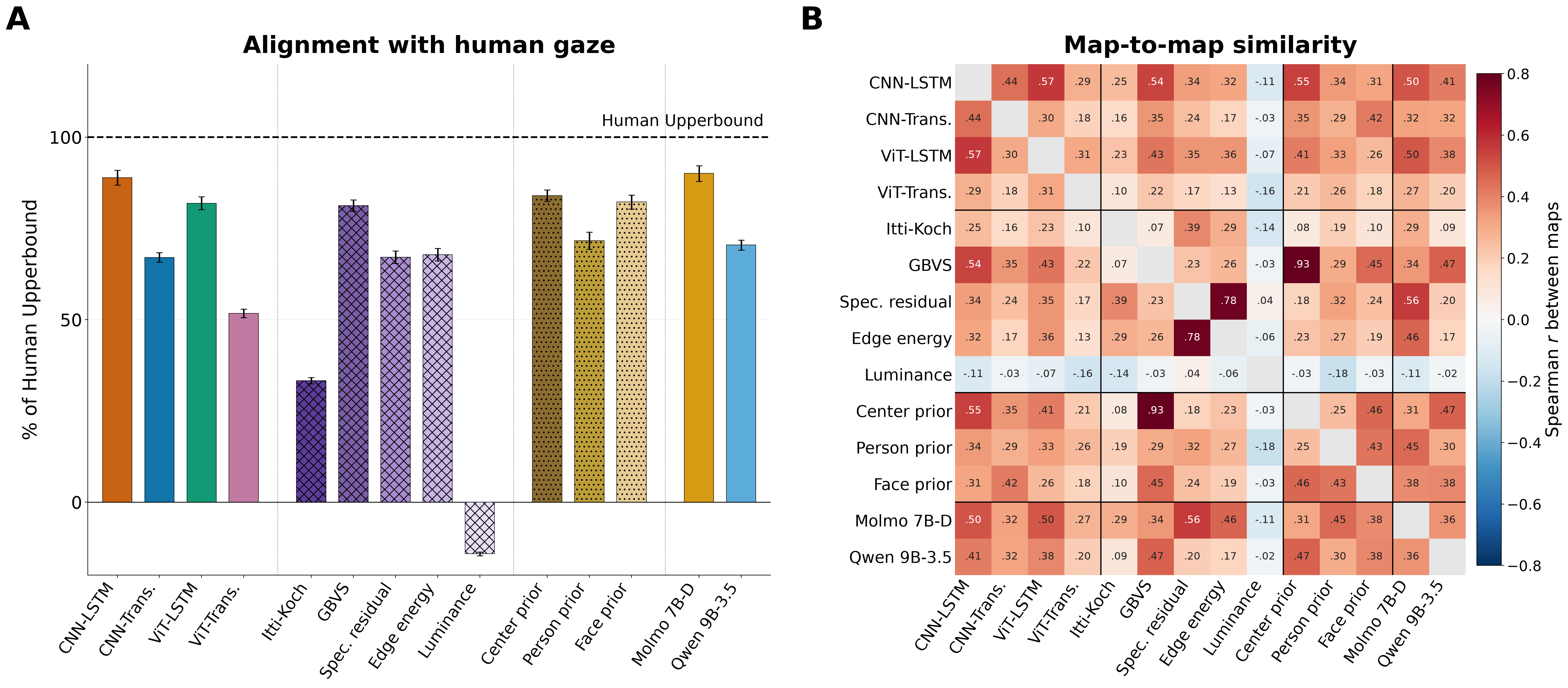}
  \caption{Baselines. \textbf{(A)} Alignment with human fixations as a percentage of the human upperbound, averaged over the Social and Describe tasks, for the four trained models, the two VLMs, and eight baselines. \textbf{(B)} Spearman correlation between the maps of every pair of predictors on each image ($14{\times}14$ grid), averaged over images and tasks.}
  \label{fig:baselines}
\end{figure}

Figure~\ref{fig:baselines}A shows the alignment of each baseline, averaged across tasks, next to our
trained models. Table~\ref{tab:baselines} reports alignment in each task. In comparing our models with image-statistic baselines, we asked whether our models predicted gaze beyond what visual statistics alone 
explained. As reported in Section~\ref{sec:results:baselines}, the most aligned behavioral model beat the most aligned saliency model, confirming, for at least one model, that this was the case. The priors were harder to beat. At the task level, the center prior reached 94\% of the upperbound in the Describe task, more than any trained model, and the person prior reached 98\% in the Social task, more than any trained model. The face prior matched CNN-LSTM in the Social task ($d_z = 0.03$, $p = .67$) and fell below it in the Describe task ($d_z = 0.25$, $p = .002$). The alignment score for Molmo was about the same as our best performing model; Qwen was much less aligned than that model, by comparison ($d_z = 0.57$ and $0.71$, both $p < .001$).

Because the center prior both predicted gaze well and resembled our best model, we asked whether the models predicted fixations beyond it. For each participant and image, we computed the partial Spearman correlation between the model map and the fixation map, controlling for the center prior, and recomputed the upperbound in the same way. All four models remained aligned with human gaze (all $p < .001$). The LSTM still outscored the Transformer sharing its encoder in all four encoder-by-task comparisons, significantly in three of them ($d_z = 0.52$ to $0.81$, all $p < .001$; CNN, Describe: $d_z = 0.14$, $p = .055$). Among the baselines, GBVS fell to about 20\%, while spectral residual, edge energy, and the person and face priors kept most of their alignment. Our models therefore carry image-specific information that the center prior cannot supply, but this advantage is smaller than the raw alignment scores suggest.

We also compared all attention maps directly, correlating every pair on each image (Figure~\ref{fig:baselines}B). The LSTM models tracked the center prior ($r = .55$ CNN-LSTM, $.41$ ViT-LSTM) more closely than the Transformer models did ($r = .35$, $.21$), while CNN-Transformer tracked faces more closely than CNN-LSTM did ($r = .42$ vs.\ $.31$). Among the saliency models, every trained model resembled GBVS most. GBVS, however, was nearly identical to the center prior ($r = .93$), so this resemblance may reflect a shared center bias rather than a shared measure of salience. Molmo's maps were the most edge-like of any model (edge energy $r = .46$, spectral residual $r = .56$), and the LSTM models were the trained models closest to Molmo ($r = .50$ for both).

\begin{table}[t]
\centering
\caption{Alignment of model attention and baselines with human fixations, before and after removing center bias. Raw: Spearman $r$ against each participant's fixation map (mean over participants and images). Partial: the same, controlling for a leave-one-image-out center prior. Percentages are of the human upperbound computed under the same metric (raw: $r = 0.272$ Social, $0.248$ Describe; partial: $0.237$, $0.203$). Rows are grouped as in Figure~\ref{fig:baselines}A. From top to bottom, the groups are our four trained models, the image-statistic baselines, the priors, and the two pretrained VLMs. Transformer and VLM rows use the cross-validated per-task head. Saliency, low-level, and semantic-prior baselines are reported at their own best Gaussian blur. Edge energy and luminance are computed in each cell of the $14{\times}14$ grid, as in the neural analysis. The two semantic priors use ground-truth annotation from the COCO dataset.}
\label{tab:baselines}
\small
\setlength{\tabcolsep}{4.5pt}
\begin{tabular}{lrrrrrrrr}
\toprule
 & \multicolumn{4}{c}{Social} & \multicolumn{4}{c}{Describe} \\
\cmidrule(lr){2-5} \cmidrule(lr){6-9}
 & raw $r$ & \% & partial $r$ & \% & raw $r$ & \% & partial $r$ & \% \\
\midrule
CNN-LSTM & 0.241 & 89 & 0.161 & 68 & 0.222 & 89 & 0.109 & 54 \\
CNN-Transformer & 0.201 & 74 & 0.128 & 54 & 0.152 & 61 & 0.098 & 48 \\
ViT-LSTM & 0.221 & 81 & 0.157 & 66 & 0.206 & 83 & 0.128 & 63 \\
ViT-Transformer & 0.155 & 57 & 0.122 & 52 & 0.118 & 48 & 0.072 & 36 \\
\midrule
Itti--Koch \citep{itti1998model} & 0.101 & 37 & 0.083 & 35 & 0.103 & 42 & 0.083 & 41 \\
GBVS \citep{harel2006graph} & 0.206 & 76 & 0.048 & 20 & 0.229 & 92 & 0.042 & 21 \\
Spectral residual \citep{hou2007saliency} & 0.186 & 68 & 0.144 & 60 & 0.191 & 77 & 0.146 & 72 \\
Edge energy (Sobel) & 0.183 & 67 & 0.126 & 53 & 0.198 & 80 & 0.134 & 66 \\
Luminance & -0.041 & -15 & -0.030 & -13 & -0.029 & -12 & -0.018 & -9 \\
\midrule
Center prior (mean fixation map) & 0.203 & 74 & -0.005 & -2 & 0.233 & 94 & -0.005 & -3 \\
Person mask (COCO annotation) & 0.266 & 98 & 0.218 & 92 & 0.200 & 80 & 0.147 & 72 \\
Face blobs (COCO keypoints) & 0.237 & 87 & 0.158 & 67 & 0.195 & 79 & 0.109 & 54 \\
\midrule
Molmo 7B-D & 0.234 & 86 & 0.185 & 78 & 0.234 & 94 & 0.180 & 89 \\
Qwen3.5 9B & 0.207 & 76 & 0.129 & 54 & 0.163 & 66 & 0.065 & 32 \\
\bottomrule
\end{tabular}
\end{table}

\section{Metric Suite}
\label{app:metric-suite}

\begin{figure}[h]
  \centering
  \includegraphics[width=\linewidth]{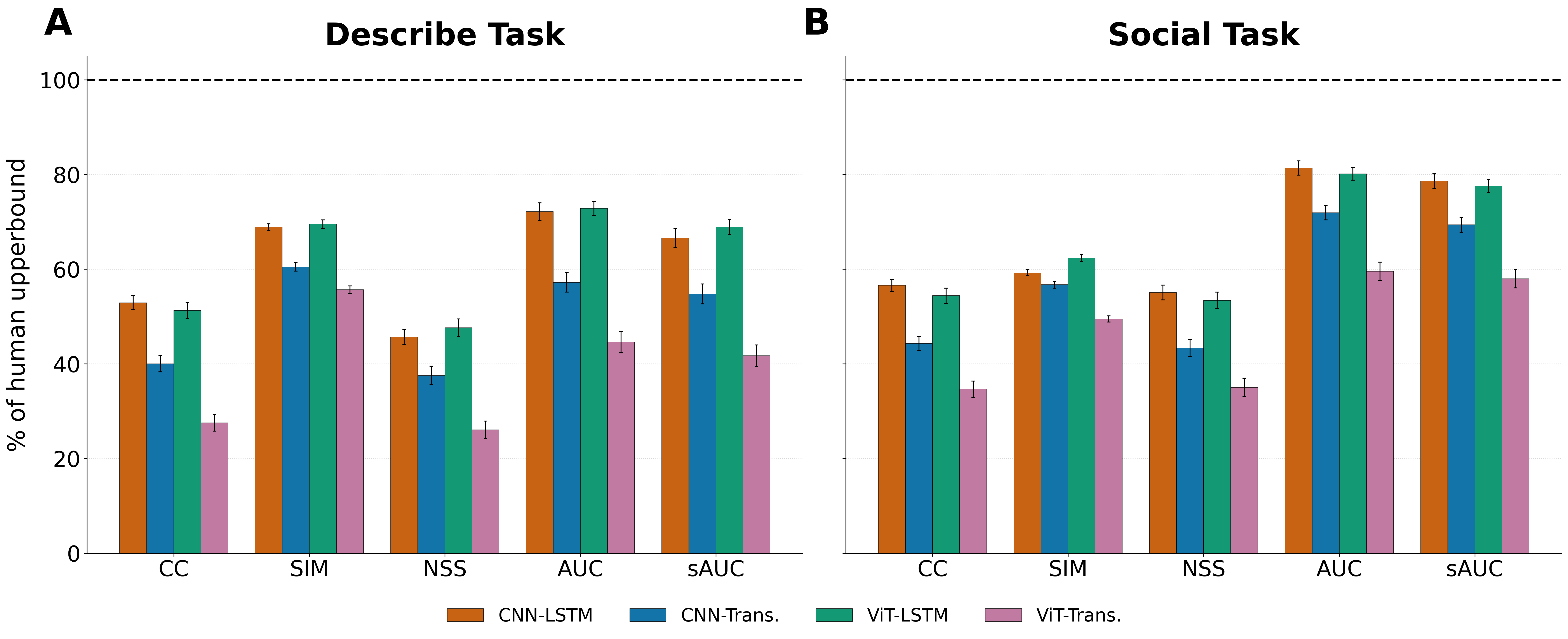}
  \caption{Scores on five additional alignment metrics for the Describe task (A) and the Social task (B). Each metric (CC, SIM, NSS, AUC-Judd, shuffled AUC) is normalized by its own split-half human upperbound (dashed line, 100\%) and shown for the four trained models on cross-validated heads. Error bars are SEM across images.}
  \label{fig:metric-suite}
\end{figure}

To make sure our results were not an artifact of the specific similarity metric we used, we compared each model's attention map for every image with that image's participant-averaged heatmap on five additional alignment metrics that are standard in the saliency literature \citep{bylinskii2019different}. Two of these metrics compare the model and human maps across the entire image. The Pearson correlation (CC) measures how closely the model map's values follow the human heatmap's values across all pixels. Histogram intersection (SIM) treats each map as a distribution of attention that sums to 1 and measures the degree to which model and participant attention overlap, from 0 (no overlap) to 1 (identical). The other three metrics compare how the model assigns attention on the most-fixated regions. We labeled the most-fixated regions as the top 5\% of pixels in each image's participant-averaged heatmap, and called these masked regions pseudo-fixations. Normalized scanpath saliency (NSS) rescales the model map to mean 0 and SD 1 and takes its average value at the pseudo-fixated pixels, so that 0 is chance. AUC-Judd is the probability that the model gives a randomly chosen pseudo-fixated pixel a higher value than a randomly chosen other pixel in the same image (0.5 is chance). Shuffled AUC (sAUC) is the same probability when the comparison pixel is drawn from the pseudo-fixated locations of all images in the task, which discounts a general bias toward the image center.

To place the metrics on a common scale, we expressed each as a percentage of its human upperbound, defined as the score obtained when the averaged heatmap of a random half of participants predicts that of the other half (mean of 20 splits). For AUC and sAUC, we subtracted chance (0.5) before dividing. 

The decoder result held under every metric. Each LSTM outscored the Transformer sharing its encoder on all five metrics in both tasks (20 of 20 contrasts, all $p < .001$ by paired permutation over images), with effects ranging from $d_z = 0.25$ (CNN, Social, SIM) to $d_z = 1.09$ (ViT, Social, SIM). The metrics also largely agreed on the ranking of the four models (mean pairwise Kendall $\tau = 0.87$ in both tasks).

\section{Smoothness}
\label{app:smoothness}

\begin{figure}[h]
  \centering
  \includegraphics[width=\linewidth]{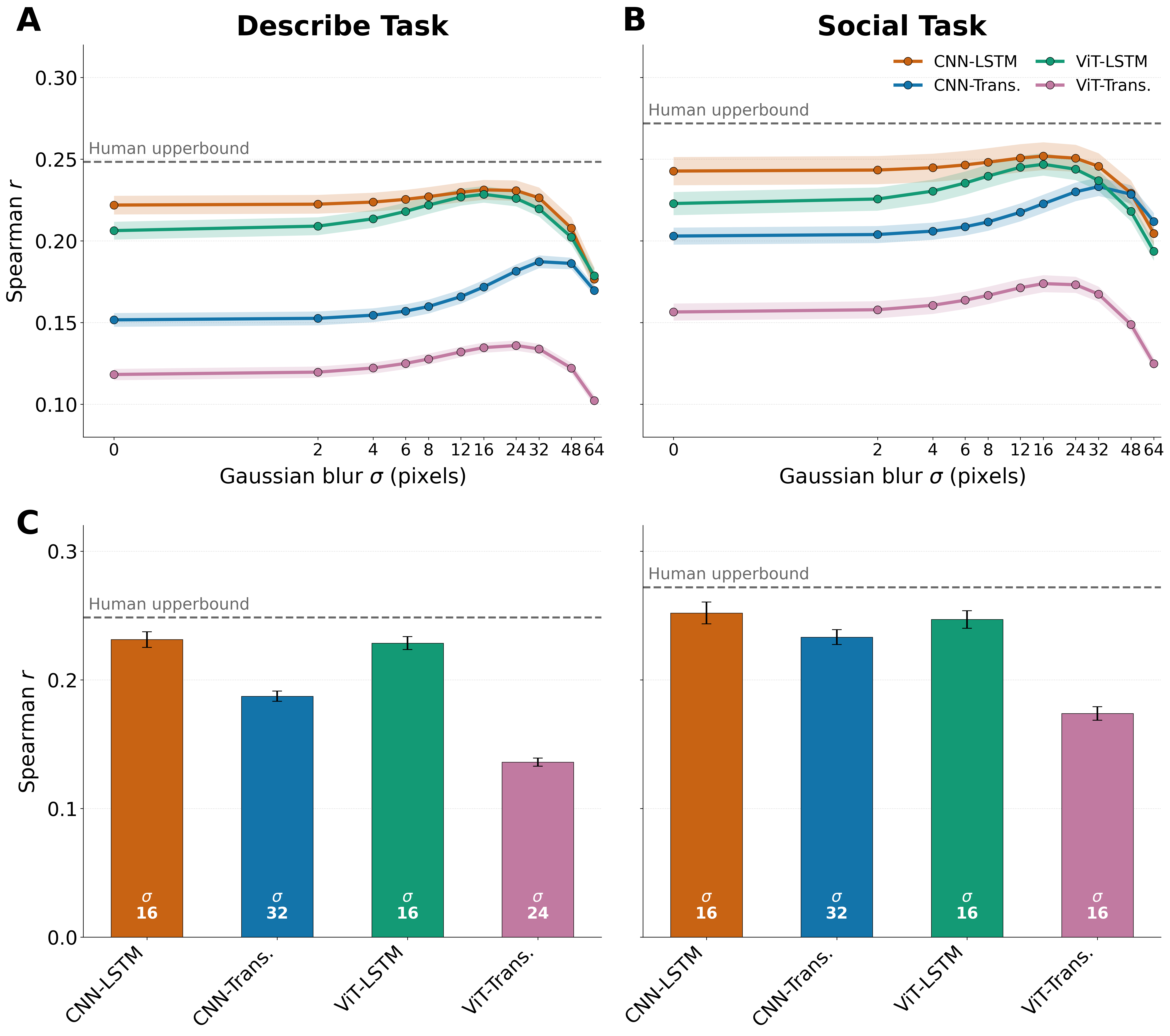}
  \caption{\textbf{(A, B)} Alignment as a function of Gaussian blur applied to each model's attention map ($\sigma$ in pixels of the $256{\times}256$ map) for the Describe (A) and Social (B) tasks. $\sigma = 0$ is the unblurred map used in the main text. Shading: SEM across participants. \textbf{(C)} Alignment at each model's cross-validated Gaussian blur for the Describe (left) and Social (right) tasks. The $\sigma$ chosen for each model is printed at the base of its bar. Error bars: SEM across participants. In all panels, the dashed line is the human upperbound.}
  \label{fig:blur-sweep}
\end{figure}

To make sure alignment did not depend on how smooth the model maps were, we blurred every model's attention map with a Gaussian kernel at eleven widths ($\sigma = 0$, 2, 4, 6, 8, 12, 16, 24, 32, 48, and 64~px on the $256{\times}256$ map). We then scored each blurred map against each participant's fixation heatmap exactly as in the main analysis. Each model's blur was chosen separately for each task by five-fold cross-validation over images: the $\sigma$ with the highest mean alignment on four folds (160 images) was scored on the held-out fold (40 images), and the five held-out scores were averaged.

Blurring raised alignment for every model, by 3.5 to 14.4 percentage points of the upperbound, but the ordering did not change at each model's best blur (Figure~\ref{fig:blur-sweep}C). In both tasks, CNN-LSTM remained the most aligned model, followed by ViT-LSTM, CNN-Transformer, and ViT-Transformer (Describe: 92.8\%, 91.7\%, 74.7\%, 53.9\%; Social: 92.3\%, 90.4\%, 85.2\%, 63.0\%). The LSTM was more aligned than the Transformer sharing its encoder in all four encoder-by-task comparisons ($d_z = 0.81$ to $6.96$ across participants, all $p_\mathrm{Holm} < .001$).

\section{Neural Encoding: Baseline VLMs and Saliency Models}
\label{app:nsd-robustness}

\begin{figure}[h]
  \centering
  \includegraphics[width=\linewidth]{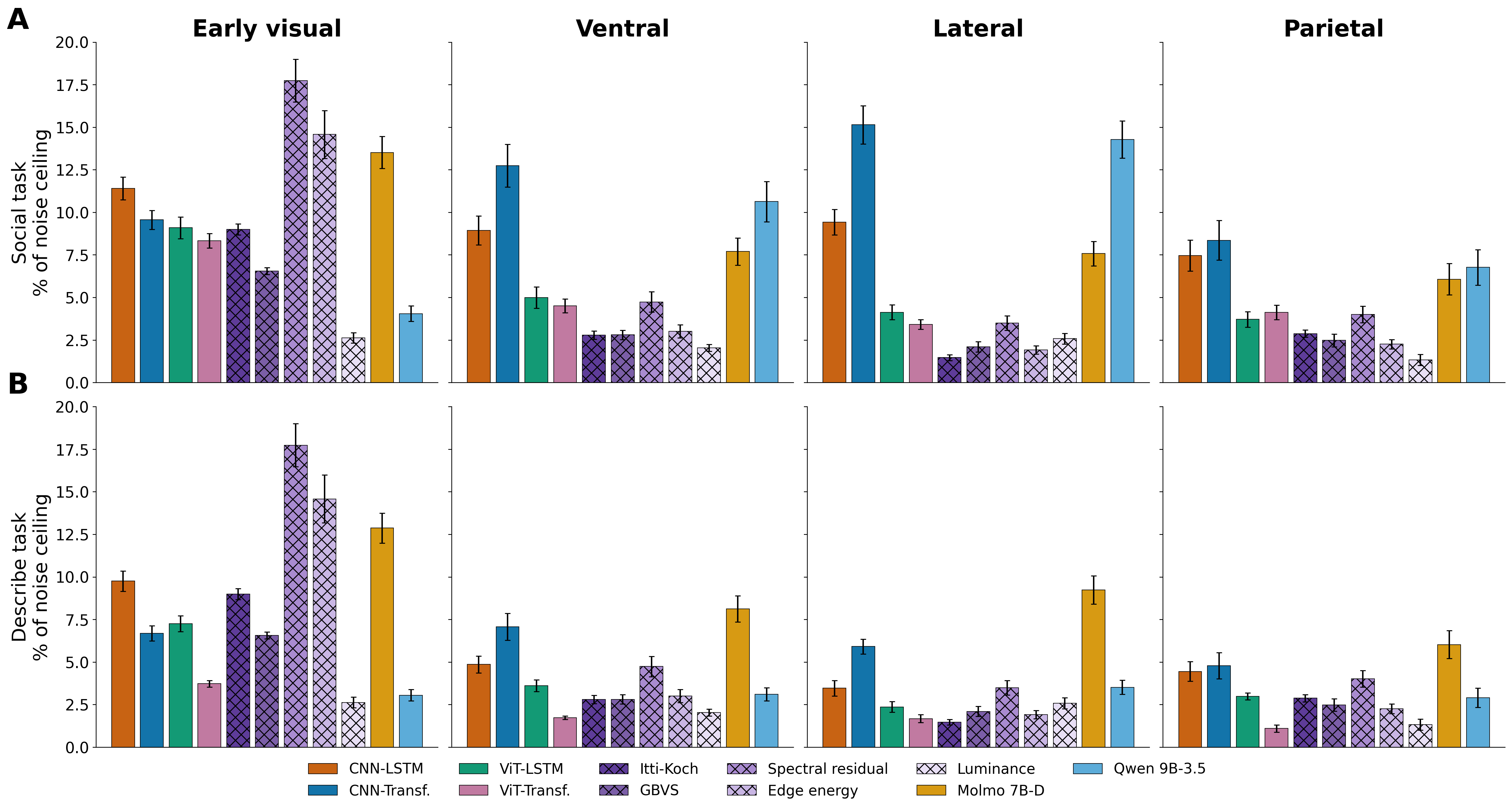}
  \caption{Neural prediction by the baseline VLMs and three bottom-up saliency models, alongside the trained models and low-level controls from Figure~\ref{fig:neural}. Held-out test $R^2$ by region in the Social (A) and Describe (B) tasks, as a percentage of each participant's noise ceiling in that region (as in Figure~\ref{fig:neural}). Saliency maps and low-level controls do not depend on the task, so their bars are identical in A and B. For the Transformer models and VLMs, each task was fit with the head selected for that task in the behavioral analysis. Error bars: SEM across the 8 participants.}
  \label{fig:nsd-vlm-saliency}
\end{figure}

We fit the attention maps of Molmo and Qwen3.5 to the NSD responses, along with maps from three bottom-up saliency models (Itti-Koch, GBVS, and spectral residual; Section~\ref{sec:methods:baselines}), using the same encoding model as the main text (Figure~\ref{fig:nsd-vlm-saliency}). To make the encoding models comparable, every predictor entered the ridge regression on the same $14{\times}14$ grid (196 features), so that differences in prediction could not reflect differences in the number of features. Molmo's $28{\times}28$ maps were area-averaged down to it, and Qwen3.5's maps ($13{\times}13$ on the NSD images) were resampled up to it. The saliency maps, computed at their own internal resolution ($32{\times}32$ for Itti-Koch and GBVS, $64{\times}64$ for spectral residual), were area-averaged down to it, as were edge energy and luminance. Every map was then normalized to sum to 1. Both VLMs and all three saliency models exceeded their own permutation floor in every region in both tasks (all $p_\mathrm{Holm} < .01$).

All three saliency models predicted early visual cortex well, but their prediction fell at higher stages of visual processing. Spectral residual predicted early visual cortex better than any other predictor (17.7\% of the noise ceiling, against 14.6\% for edge energy and 13.5\% for Molmo in the Social task). These results mirror the low-level controls in the main text: bottom-up salience accounts for responses in early visual cortex, but not for the responses in higher regions that the trained models predicted.

Molmo, the more behaviorally aligned of the two VLMs (Appendix~\ref{app:baselines}), was also among the most aligned predictors neurally. It predicted early visual cortex better than any other model in both tasks (13.5\% of the noise ceiling in the Social task, 12.9\% in the Describe task), and its prediction generally receded across later regions (parietal 6.1\% and 6.0\%). The pattern was nearly identical across tasks.

Qwen3.5 showed the opposite pattern, but only in the Social task. There it was the weakest model in early visual cortex (4.1\% of the noise ceiling), yet its prediction rose in ventral (10.6\%) and lateral cortex (14.3\%), second only to CNN-Transformer, before falling again in parietal cortex (6.8\%). In the Describe task its prediction was low and flat across regions (2.9\% to 3.5\%).

We suspect that the contrast between Molmo and Qwen3.5 reflects what each model's attention tracks. Molmo's maps were the most edge-like of any model (mean per-image Spearman correlation with edge energy $r = .45$ Social, $.38$ Describe, against $.04$ to $.29$ for the trained models) and the most concentrated, a profile suited to early visual cortex, which is sensitive to fine-grained image structure. Qwen3.5's maps were nearly unrelated to edges ($r = .03$ Social, $.06$ Describe). In the Social task, where it is important to find specific people and the objects of their attention, Qwen3.5 may track those people and objects rather than local image structure. Such a map would be of little use to early visual cortex but would help predict ventral and lateral regions, which respond to objects, faces, and bodies.

\section*{NeurIPS Paper Checklist}

\begin{enumerate}

\item {\bf Claims}
    \item[] Question: Do the main claims made in the abstract and introduction accurately reflect the paper's contributions and scope?
    \item[] Answer: \answerYes{}
    \item[] Justification: The three contributions listed in the introduction (decoder architecture as the dominant predictor of behavioral alignment, the trade-off between where attention falls and how sharply it is concentrated, and the encoding of NSD responses by attention weights, driven by the encoder) are each supported by results reported in Section~4.

\item {\bf Limitations}
    \item[] Question: Does the paper discuss the limitations of the work performed by the authors?
    \item[] Answer: \answerYes{}
    \item[] Justification: A limitations paragraph at the end of Section~5 identifies two constraints: (1) models and participants did not view the images in the same way, since people explore a new image with successive fixations while a model reads all 196 locations at once, which may favor diffuse maps; and (2) the architecture comparison is not fully controlled, since decoder type is entangled with decoder size and depth.

\item {\bf Theory assumptions and proofs}
    \item[] Question: For each theoretical result, does the paper provide the full set of assumptions and a complete (and correct) proof?
    \item[] Answer: \answerNA{}
    \item[] Justification: The paper presents no theoretical results; all contributions are empirical.

    \item {\bf Experimental result reproducibility}
    \item[] Question: Does the paper fully disclose all the information needed to reproduce the main experimental results of the paper to the extent that it affects the main claims and/or conclusions of the paper (regardless of whether the code and data are provided or not)?
    \item[] Answer: \answerYes{}
    \item[] Justification: Architecture details are given in Table~1 and Section~3.2; the three-phase training curriculum and all hyperparameters are in Section~\ref{sec:methods:training}; attention extraction procedures are in Section~\ref{sec:methods:attention}; evaluation metrics are in Section~\ref{sec:methods:stats}. COCO images are publicly available.

\item {\bf Open access to data and code}
    \item[] Question: Does the paper provide open access to the data and code, with sufficient instructions to faithfully reproduce the main experimental results, as described in supplemental material?
    \item[] Answer: \answerYes{}
    \item[] Justification: Training and analysis code and the eye-tracking fixation heatmaps are public at \url{https://github.com/isaacrc/vlm-attention-alignment}, and the trained model checkpoints at \url{https://huggingface.co/isaacrc/vlm-attention-alignment}. COCO images and the Natural Scenes Dataset are publicly available from their providers.

\item {\bf Experimental setting/details}
    \item[] Question: Does the paper specify all the training and test details (e.g., data splits, hyperparameters, how they were chosen, type of optimizer) necessary to understand the results?
    \item[] Answer: \answerYes{}
    \item[] Justification: Optimizer (Adam), learning rates, batch sizes, dropout, beam search, encoder unfreezing strategy, stopping criteria, training phases, and data splits (Karpathy splits) are reported in Section~\ref{sec:methods:training}. Attention head selection procedure is in Section~\ref{sec:methods:attention} and Appendix~\ref{app:head-selection}.

\item {\bf Experiment statistical significance}
    \item[] Question: Does the paper report error bars suitably and correctly defined or other appropriate information about the statistical significance of the experiments?
    \item[] Answer: \answerYes{}
    \item[] Justification: Architecture contrasts in behavioral alignment use paired sign-flip permutation tests over the 200 images (10,000 permutations) with Cohen's $d_z$ and Holm correction, supported by a linear mixed model (Section~\ref{sec:methods:stats}). Neural encoding results are tested against each predictor's permutation floor with Holm-corrected paired $t$-tests and a repeated-measures ANOVA with Greenhouse-Geisser correction across the 8 NSD participants (Section~\ref{sec:methods:neural}). Error bars are standard errors of the mean throughout.

\item {\bf Experiments compute resources}
    \item[] Question: For each experiment, does the paper provide sufficient information on the computer resources (type of compute workers, memory, time of execution) needed to reproduce the experiments?
    \item[] Answer: \answerYes{}
    \item[] Justification: All four architecture variants were trained on a single NVIDIA A100 GPU (40~GB, PCIe) on Princeton University's Della cluster. Each model required approximately 25 GPU-hours for the full three-phase training pipeline (Phases 1--3); total training compute was approximately 100 GPU-hours across the four variants.

\item {\bf Code of ethics}
    \item[] Question: Does the research conducted in the paper conform, in every respect, with the NeurIPS Code of Ethics \url{https://neurips.cc/public/EthicsGuidelines}?
    \item[] Answer: \answerYes{}
    \item[] Justification: Human subjects research was approved by an Institutional Review Board, participants provided written informed consent and were compensated, and no harmful content is generated or released.

\item {\bf Broader impacts}
    \item[] Question: Does the paper discuss both potential positive societal impacts and negative societal impacts of the work performed?
    \item[] Answer: \answerNA{}
    \item[] Justification: This is foundational cognitive science research characterizing how architectural choices affect attention alignment between VLMs and humans. The models produced are small image-captioning systems with no identified path to negative societal application.

\item {\bf Safeguards}
    \item[] Question: Does the paper describe safeguards that have been put in place for responsible release of data or models that have a high risk for misuse (e.g., pre-trained language models, image generators, or scraped datasets)?
    \item[] Answer: \answerNA{}
    \item[] Justification: The trained models are small image-captioning systems (LSTM and Transformer decoders on COCO). They pose no identified misuse risk and do not fall into high-risk categories such as large generative models or scraped datasets.

\item {\bf Licenses for existing assets}
    \item[] Question: Are the creators or original owners of assets (e.g., code, data, models), used in the paper, properly credited and are the license and terms of use explicitly mentioned and properly respected?
    \item[] Answer: \answerYes{}
    \item[] Justification: All assets are cited: COCO 2014 (CC BY 4.0) \citep{lin2014microsoft}, ResNet-101 \citep{he2016deep}, ViT-B/16 \citep{dosovitskiy2021image}, Molmo 7B-D \citep{deitke2024molmo}, Qwen3.5 9B \citep{qwen35team2025}, and the Natural Scenes Dataset \citep{allen2022massive}, used under its terms of use.

\item {\bf New assets}
    \item[] Question: Are new assets introduced in the paper well documented and is the documentation provided alongside the assets?
    \item[] Answer: \answerYes{}
    \item[] Justification: The eye-tracking fixation heatmaps, code and trained checkpoints are released with documentation (README with usage instructions) at \url{https://github.com/isaacrc/vlm-attention-alignment} and \url{https://huggingface.co/isaacrc/vlm-attention-alignment}.

\item {\bf Crowdsourcing and research with human subjects}
    \item[] Question: For crowdsourcing experiments and research with human subjects, does the paper include the full text of instructions given to participants and screenshots, if applicable, as well as details about compensation (if any)?
    \item[] Answer: \answerYes{}
    \item[] Justification: Task instructions are described in Section~3.1 (participants described the scene or verbally identified the attentional target of a person in the image). Participants were undergraduate students compensated with course credit. No crowdsourcing was used; all data were collected in a controlled laboratory setting.

\item {\bf Institutional review board (IRB) approvals or equivalent for research with human subjects}
    \item[] Question: Does the paper describe potential risks incurred by study participants, whether such risks were disclosed to the subjects, and whether Institutional Review Board (IRB) approvals (or an equivalent approval/review based on the requirements of your country or institution) were obtained?
    \item[] Answer: \answerYes{}
    \item[] Justification: IRB approval is stated in Section~3.1. Participants provided written informed consent prior to participation. The study involved viewing images and describing them aloud; risks were minimal and disclosed.

\item {\bf Declaration of LLM usage}
    \item[] Question: Does the paper describe the usage of LLMs if it is an important, original, or non-standard component of the core methods in this research? Note that if the LLM is used only for writing, editing, or formatting purposes and does \emph{not} impact the core methodology, scientific rigor, or originality of the research, declaration is not required.
    \item[] Answer: \answerNA{}
    \item[] Justification: Large VLMs (Molmo, Qwen3.5) serve as baselines in the study rather than as methodology tools. They were not used to develop the analysis pipeline or generate scientific claims.

\end{enumerate}

\end{document}